%% file: main.tex
\documentclass{vgtc}                          

\ifpdf
  \pdfoutput=1\relax                   
  \usepackage{graphicx}                
  \DeclareGraphicsExtensions{.pdf,.png,.jpg,.jpeg} 
\else
  \usepackage{graphicx}                
  \DeclareGraphicsExtensions{.eps}     
\fi%

\graphicspath{{figures/}{pictures/}{images/}{./}} 

\usepackage{microtype}                 
\PassOptionsToPackage{warn}{textcomp}  
\usepackage{textcomp}                  
\usepackage{mathptmx}                  
\usepackage{times}                     
\usepackage{cite}                      
\usepackage{tabu}                      
\usepackage{booktabs}                  
\usepackage{multicol}
\usepackage{multirow}

\usepackage{subfiles}
\usepackage{xcolor}
\usepackage{graphics}

\newcommand{\rev}[1]{#1}

\onlineid{0}

\vgtccategory{Research}

\vgtcinsertpkg

\title{Uniform Manifold Approximation with Two-phase Optimization}

\author{
    Hyeon Jeon ${}^1$ \thanks{equal contribution}$\:\:\:$ \thanks{e-mail: hj@hcil.snu.ac.kr}  $\quad$ 
    Hyung-Kwon Ko ${}^2$ ${}^*$ \thanks{e-mail: hkko@hcil.snu.ac.kr} $\quad$
    Soohyun Lee ${}^1$ \thanks{e-mail: shlee@hcil.snu.ac.kr} $\quad$
    Jaemin Jo ${}^3$ \thanks{e-mail: jmjo@skku.edu} $\quad$
    Jinwook Seo ${}^1$ \thanks{e-mail: jseo@snu.ac.kr} \\
    \scriptsize ${}^1$ Seoul National University \\
    \scriptsize ${}^2$ NAVER Webtoon Corp. \\
    \scriptsize ${}^3$ Sungkyunkwan University \\
}

\abstract{
We introduce Uniform Manifold Approximation with Two-phase Optimization (UMATO), a dimensionality reduction (DR) technique that improves UMAP to capture the global structure of high-dimensional data more accurately. In UMATO, optimization is divided into two phases so that the resulting embeddings can depict the global structure reliably while preserving the local structure with sufficient accuracy. In the first phase, hub points are identified and projected to construct a skeletal layout for the global structure. In the second phase, the remaining points are added to the embedding preserving the regional characteristics of local areas. Through quantitative experiments, we found that UMATO (1) outperformed widely used DR techniques in preserving the global structure while (2) producing competitive accuracy in representing the local structure. We also verified that UMATO is preferable in terms of robustness over diverse initialization methods, numbers of epochs, and subsampling techniques.
} 

\CCScatlist{
  \CCScatTwelve{Human-centered computing}{Visualization}{Visualization techniques}{};
  \CCScatTwelve{Computing methodologies}{Machine learning}{Machine learning algorithms}{}
}

\begin{document}


\firstsection{Introduction}

\maketitle

\subfile{sections/01-introduction}

\subfile{sections/02-related-works}

\subfile{sections/03-umato}

\subfile{sections/04-experiments}

\subfile{sections/05-additional-experiments}
\subfile{sections/06-conclusion}

\acknowledgments{
This work was supported by the National Research Foundation of Korea (NRF) grant funded by the Korea government (MSIT) (No. NRF2019R1A2C208906213)}

\bibliographystyle{abbrv-doi}

\bibliography{ref}
\end{document}

%% file: sections/01-introduction.tex
Dimensionality reduction (DR) is one of the most useful tools for exploring high-dimensional (HD) data in visual analytics \cite{jo2018panene, fujiwara2019supporting, chatzimparmpas2020t}. In various domains, DR provides an effective means of understanding HD data by enabling visual inspection of HD data \cite{liu2016visualizing, kehrer2012visualization, tang2016visualizing}. Especially, nonlinear DR techniques such as $t$-SNE \cite{maaten2008visualizing}, NeRV \cite{venna2010information}, and UMAP \cite{mcinnes2018umap} made huge success in capturing complex local manifolds \cite{lee2014two}. 

However, despite the success, nonlinear DR techniques often work poor in preserving the global structure of data \cite{fu2019atsne, kobak2021initialization}. For example, in the case of $t$-SNE, inaccuracy in preserving the global structure comes from the fact that its loss function, Kullback-Leibler (KL) divergence, assigns too little penalty for the points that are distant in the original space and stay close in the embedding space \cite{mcinnes2018umap}. 
UMAP avoids such a problem by employing the cross-entropy function as a loss function. However, it still barely captures the global structure as it optimizes the layout based on a $k$-nearest neighbor ($k$NN) graph. UMAP's approximation for fast computation (\autoref{sec:umap}) also leads to inaccuracy; for example, a recent study \cite{kobak2019umap} showed that UMAP does not outperform $t$-SNE in preserving the global structure given the same initial layout. One way to alleviate this problem is to not use approximation, which would increase the computational cost greatly.

To overcome the aforementioned limitations, we present Uniform Manifold Approximation with Two-phase Optimization (UMATO), a novel DR technique that enhances UMAP to better preserve the global structure. In UMATO, optimization is first run for a small number of points that represent the data (i.e., hub points). As finding the optimal embedding for a small number of points is relatively easy and robust, this makes approximation unnecessary and thus leads to the better preservation of the global structure. Moreover, this procedure makes the embedding more stable and less sensitive to initialization methods. After capturing the overall skeleton of the HD structure with the hub points, we gradually append the rest of the points in subsequent phases. Although the same approximation techniques as UMAP are used for these points, the 
embedding becomes more accurate in preserving the global structure because we use already embedded hub points as anchors.

We quantitatively compared UMATO against five widely used DR techniques using four datasets. The embeddings were then analyzed with global and local quality metrics such as KL divergence and Trustworthiness \& Continuity \cite{venna2001neighborhood}, respectively. As a result, we found that UMATO outperformed the competitors in preserving the global structure while showing competitive performance in preserving the local structure. We also found that UMATO is more stable against sampling fluctuations and changes in the number of epochs, and is less sensitive to the initialization methods, compared to the competitors (Appendix D, E, and F).

%% file: sections/02-related-works.tex
\section{Background and Related Works}

\subsection{UMAP}

\label{sec:umap}

Because UMATO shares the overall pipeline of UMAP, we provide a brief introduction to UMAP. Although UMAP is grounded in a sophisticated mathematical foundation, its computation can be simply divided into two steps, graph construction and layout optimization.

\noindent \textbf{Graph Construction.} UMAP starts by generating a weighted $k$NN graph; given $k$ (the number of NN to consider) and a distance metric $d: X \times X \rightarrow [0, \infty)$, UMAP first computes $\mathcal{N}_i$, the $k$NN of $x_i$ with respect to $d$.
Then, two parameters, $\rho_i$ and $\sigma_i$, are computed for each data point $x_i$ to identify its local metric space.
$\rho_i$ is a nonzero distance from $x_i$ to its nearest neighbor:

\begin{equation} \label{eq:rho}
    \rho_{i} = \min_{j \in \mathcal{N}_i}\{ d(x_{i}, x_j) \ | \ d(x_{i}, x_j) > 0 \}.
\end{equation}

Using binary search, UMAP finds $\sigma_i$  that satisfies:

\begin{equation} \label{eq:sigma}
    \sum_{j \in \mathcal{N}_i} \exp({-\max(0, d(x_{i}, x_j) - \rho_{i})} / \sigma_{i}) = \log_2(k).
\end{equation}

Next, UMAP computes:

\begin{equation} \label{eq:v}
    v_{j|i} = \exp({-\max(0, d(x_{i},x_{j})-\rho_{i})} / \sigma_{i}),
\end{equation}
which is the weight of the edge from a point $x_i$ to another point $x_j$. A combined weight of a edge is then defined as $v_{ij} = v_{j|i} + v_{i|j} - v_{j|i} \cdot v_{i|j}$.
In the embedded space, the similarity between two points $y_i$ and $y_j$ is defined as $w_{ij} = (1 + a|| y_{i} - y_{j} ||_{2}^{2b})^{-1}$, where $a$ and $b$ are positive constants defined by users. 
Setting both $a$ and $b$ to 1 is identical to using Student's $t$-distribution to measure the similarity.

\noindent \textbf{Layout Optimization.}
The goal of layout optimization is to find the $y_i$ that minimizes the difference (or loss) between $v_{ij}$ and $w_{ij}$.
Unlike $t$-SNE, UMAP employs cross-entropy:
\begin{equation} \label{eq:ce}
    CE = \sum_{i \neq j} [v_{ij} \cdot \log({v_{ij}} / {w_{ij}}) - (1-v_{ij}) \cdot \log((1-v_{ij}) / (1-w_{ij}))],
\end{equation}
between $v_{ij}$ and $w_{ij}$ as the loss function.
UMAP initializes $y_i$ through spectral embedding~\cite{belkin2002laplacian} and iteratively optimize\rev{s} its position to minimize $CE$.
Given the output weight $w_{ij}$ as $1/(1+ad_{ij}^{2b})$, \rev{where $d_{ij}^{2b} = || y_i = y_j ||_{2}^{2b}$}, the attractive gradient is:
\begin{equation} \label{eq:pos}
    \frac{CE}{y_i} ^{+} = \frac{-2abd_{ij}^{2(b-1)}}{1+ad_{ij}^{2b}} v_{ij} (y_i - y_j),
\end{equation}
and the repulsive gradient is:
\begin{equation} \label{eq:neg}
    \frac{CE}{y_i} ^{-} = \frac{2b}{(\epsilon + d_{ij}^{2})(1 + ad_{ij}^{2b})} (1 - v_{ij}) (y_i - y_j).
\end{equation}
$\epsilon$ is a small constant added to prevent division by zero and $d_{ij}$ is \rev{the} Euclidean distance between $y_i$ and $y_j$.
For efficient optimization, UMAP leverages the negative sampling technique \cite{mikolov2013distributed, tang2015line, tang2016visualizing}.
The technique first chooses a target edge $(i, j)$ and $M$ negative sample points. Then, $i$ and $j$ are used to compute attractive forces, while negative samples are used to compute repulsive forces; for each epoch, positions of $i$, $j$, and negative samples are updated. 
Considering negative sampling, the modified objective function is:
\begin{equation}    \label{eq:eq9}
    O = \sum_{(i, j) \in E} v_{ij} (\log (w_{ij}) + \sum_{k=1}^{M} E_{{j_k} \sim P_{n}(j)} \gamma \log (1 - w_{ij_{k}})).
\end{equation}
Here, $v_{ij}$ and $w_{ij}$ are the similarities in the high and low-dimensional spaces, respectively, and $\gamma$ is a weight constant to apply to negative samples. 
$E_{{j_k} \sim  P_n(j)}$ means that $j_k$ is sampled from a noisy distribution $P_n(j) \propto deg_j^{3/4}$ \cite{mikolov2013distributed}, where $deg_j$ denotes the degree of point $j$.

\subsection{Dimensionality Reduction Techniques for Preserving Global Structure}

\label{sec:gldr}

Preserving global structures in DR has long been considered an important research topic \cite{jeon2022measuring, fu2019atsne, Moor19Topological, kobak2021initialization, kobak2019umap}. 
One way is to design a loss function or an optimization pipeline that targets global aspects of data. 
For instance, Isomap \cite{tenenbaum2000global} preserves global structure by approximating its geodesic distances in the embedding. 
Another example is TopoAE \cite{Moor19Topological}, a deep-learning approach that uses a generative model. To make the latent space resemble the HD space, it appends a topological loss to the original reconstruction loss of autoencoders \cite{hinton2006reducing}. However, these techniques leverage a single-step optimization; thus, their focus on global structure reduces their attention to local structure. In contrast, UMATO focuses on both global and local structures by dividing the optimization into two phases.

Another common scheme to capture global structure is to utilize sample points to better model the original space; these points are usually called hubs, landmarks, or anchors. 
Silva et al. \cite{silva2003global} proposed L-Isomap, a landmark version of classical multidimensional scaling (MDS), to alleviate its computation cost. However, in L-Isomap, landmarks are chosen randomly without considering their importance. 
HSNE \cite{pezzotti2016hierarchical} and HUMAP \cite{marcilio2021humap} instead chose landmarks based on $k$NN graph, which is similar to UMATO. However, in HSNE and HUMAP, landmarks' main role is constructing the early stage of hierarchical embeddings in which users can interactively drill down. Instead, UMATO utilizes hubs to capture the global structure and work as global anchors for the layout of non-hub points.

The most similar work to ours is A$t$-SNE~\cite{fu2019atsne}, which optimizes the anchor points and remaining points with two different loss functions.
Nonetheless, as the anchors wander during the optimization and the loss function (KL divergence) does not care about distant points, it barely captures the global structure (\autoref{tab:quant}).
UMATO avoids this problem by separating hubs and non-hub points;  the hubs take their position in the first phase and barely move but guide other points in the second phase so that the global structure can be preserved robustly.
Applying cross-entropy as a loss function also helps in preserving both structures \cite{mcinnes2018umap}.

%% file: sections/03-umato.tex
\begin{figure*}[t]
    \centering
    \includegraphics[width=\linewidth]{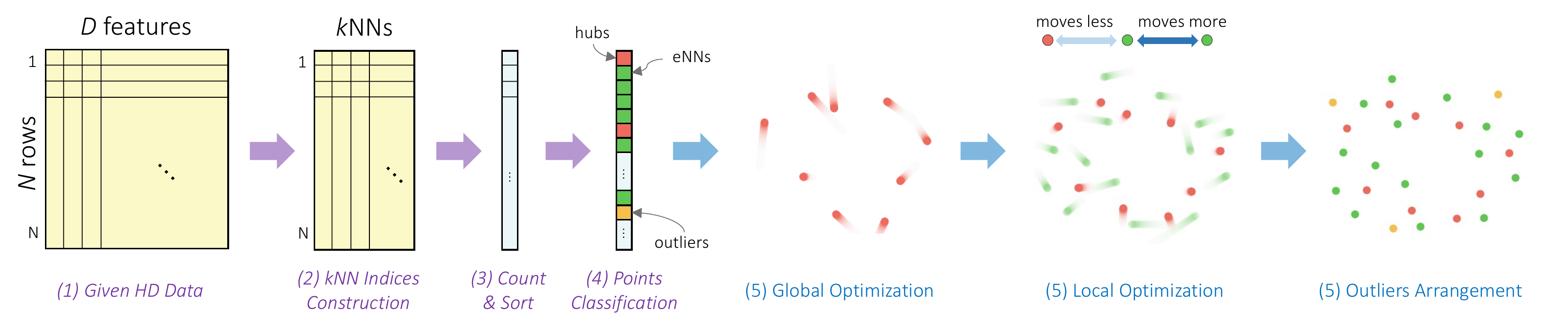}
    \vspace{-8mm}
    \caption{The illustration of the overall UMATO pipeline.
            Based on (1) given HD data,
            (2) we first find $k$NN for each point.
            (3) Then, we count the frequency of each point in the $k$NN index table and sort them in decreasing order
            (4) Next, points are categorized into hub points, expanded nearest neighbors (eNN), and outliers based on their connectivity to other points.
            (5) We initialize and optimize the positions of hubs (global optimization).
            (6) Leveraging the hubs' positions, the positions of eNN are optimized (local optimization). 
            (7) Finally, outliers are arranged. The procedures described in \autoref{sec:ptclassi} are depicted with purple arrow and captions, while the ones described in \autoref{sec:layout} are represented in blue.}
    \label{fig:illust}
    \vspace{-5.5mm}
\end{figure*}

\section{UMATO}

We present UMATO, which splits UMAP's optimization into two phases to \rev{ preserve global structure better while maintaining the capability to represent local structure}. 
For ease of understanding, we illustrated the UMATO pipeline in \autoref{fig:illust}, presented the pseudo code in Appendix A, and publicized the source code \cite{umatogithub}.

\subsection{Points Classification} 

\label{sec:ptclassi}

In overall, UMATO follows the pipeline of UMAP; 
we first find the $k$NN, and by calculating  $\rho$ (\autoref{eq:rho}) and $\sigma$ (\autoref{eq:sigma}) for each point, we obtain the pairwise similarity for every pair of points in $k$NN indices.
Once $k$NN indices are established, we unfold them and check the frequency of each point to sort them in descending order so that the index of the popular points can come to the front.

Then, we divide points into three disjoint sets: hubs, expanded nearest neighbors (eNN), and outliers. At first, we repeat the following steps until no points remain unconnected: (1) choose the point with the biggest frequency as a hub among points that are not yet selected; (2) \rev{remove} the $k$NN of the chosen hub \rev{from the sorted list}. 
Thanks to the sorted indices, the most popular point in each iteration---but not too densely located---becomes the hub point. 
Then,  we recursively seek out hub points' NN and again look for the NN of those neighbors until there are no points to be newly appended. In other words, we find all connected points that are expanded from the original hub points, which, in turn, is called eNN.
Any remaining point, neither a hub nor an eNN, becomes an outlier. 

Our optimization takes distinctive approaches for different sets so that both global and local structures can be well preserved. That is, we first optimize hubs for the global structure, then perform local optimization for eNN. 
\rev{We do not optimize outliers to prevent them from disturbing the overall structure of the embedding.}
In the following section, we explain each step in detail.

\subsection{Layout Optimization}

\label{sec:layout}

\noindent \textbf{Global Optimization. } 
At first, we run the global optimization for hubs to construct the skeletal layout of the embedding.
First, we initialize the positions of hub points using PCA, which makes the optimization process more stable than using random initial positions \cite{kobak2021initialization}.
Next, we optimize the positions of hub points by minimizing the cross-entropy (\autoref{eq:ce}).
Let $f(X) = \{ f(x_{i}, x_{j}) | x_{i}, x_{j} \in X \}$ and $g(Y) = \{ g(y_{i}, y_{j}) | y_{i}, y_{j} \in Y \}$ be two adjacency matrices in high- and low-dimensional spaces.
If $X_{h}$ represents a set of points selected as hubs in HD space and $Y_{h}$ is a set of corresponding points in the embedding, $CE(f(X_{h}) || g(Y_{h}))$ is minimized.
Unlike UMAP, our global optimization excludes negative sampling approximation; therefore, the embedding becomes more robust and less biased in capturing global structure. As the optimization runs for a few hub points, it still does not require too much time to compute.

\noindent \textbf{Local Optimization. }
In the second phase, UMATO additionally embeds eNN. 
For each eNN, we set its initial position as an average position of $m$ (e.g., 10) NN with a small random perturbation.  
We then perform local optimization, similar to UMAP but slightly different. As explained in \autoref{sec:umap}, UMAP starts by constructing a $k$NN graph; we perform the same task but only with the hubs and eNN. To do this, we need to update $k$NN indices constructed in advance (\autoref{sec:ptclassi}) to exclude outliers. We do this by simply replacing the outliers with new NN; thus, the computation is inexpensive. 

Afterward, similar to UMAP, local optimizations of hubs and eNN are performed based on the cross-entropy loss function. The negative sampling technique (\autoref{eq:eq9}) is also leveraged. However, rather than updating the positions of entire points equally, UMATO tries to keep the positions of hubs as much as possible because they have already formed the global structure. For this aim, while sampling a target edge $(i, j)$, we select $i$ among eNN and choose $j$ from both hubs and eNN. Then, if $j$ is a hub, we penalize the attractive force for $j$ by assigning a small weight (e.g., 0.1). This makes $j$ not excessively affected by $i$ if it is a hub point.
In addition, as repulsive force can disperse the local attachment, making the point veer off for each epoch and eventually destroying the well-shaped global layout, we multiply a penalty (e.g., 0.1) when calculating the repulsive gradient (\autoref{eq:neg}) for the points selected as negative samples.

\noindent \textbf{Outliers Arrangement.}
Isolated points (i.e., outliers) mostly have the same distance to all the other data points in HD space due to the curse of dimensionality \cite{bellman1966dynamic}. Therefore, optimizing outliers can make them mingle with the already embedded points (i.e., hubs, eNN), thus sabotaging global and local structures. Therefore, we do not optimize outliers but simply append them using a previously embedded point that is located nearest to each outlier in the HD space. 
That is, for an outlier $x_i \in C_n$ where $C_n$ is the connected component to which $x_i$ belongs, we first find $x_j \in C_n$ that has already been embedded and is closet to $x_i$. Then, we set the low-dimensional position of $x_i$ as the one of $x_j$ with random noise. 
This helps us benefit from the comprehensive composition of the embedding that we have already optimized.
Note that all outliers can find a previously embedded point as their neighbor. This is because we picked more than one hub from each connected component of the NN graph, and thus at least one point of each component has an optimized position (\autoref{sec:ptclassi}).

%% file: sections/04-experiments.tex
\begin{table*}[t]

\begin{center}
\small

\begin{tabular}{cccccccc|cccc}
\toprule
     &  & \multicolumn{6}{c}{Global quality metrics} & \multicolumn{4}{c}{Local quality metrics ($k$ = 5)} \\
\midrule
    Dataset & Algorithm & DTM$_{0.01}$ & DTM$_{0.1}$ & DTM$_{1}$ & KL$_{0.01}$ & KL$_{0.1}$ & KL$_{1}$ & Conti. & Trust. & MRRE$_{X}$ & MRRE$_{Z}$ \\
\midrule
    \multirow{7}{*}{Spheres} & PCA & 1.0123 & 0.9950 & 0.1687 & 0.7568 & 0.6525 & 0.0153 & 0.7983 & 0.6088 & 0.7985 & 0.6078\\
    & Isomap & \underline{0.7020} & 0.7784 & 0.1282 & 0.4492 & 0.4267 & 0.0095 & \textbf{\underline{0.9041}} & 0.6266 & \textbf{0.9039} & 0.6268\\
    & $t$-SNE & 0.9331 &0.9144& 0.1532 & 0.6091 & 0.5399 & 0.0130 &  \textbf{0.8916} & \textbf{\underline{0.7078}} & \textbf{\underline{0.9045}} & \textbf{\underline{0.7241}}\\
    & UMAP & 0.9474 & 0.9209 & 0.1548 & 0.6100 & 0.5383 & 0.0134 &   0.8760 & 0.6499 & 0.8805 & 0.6494\\
    & TopoAE & \textbf{0.4099} &\textbf{0.6890}& \textbf{0.1197} & \textbf{0.2063} & \textbf{0.3340} & \textbf{0.0076} &  0.8317 & 0.6339 & 0.8317 & 0.6326\\
    & A$t$-SNE & 0.9634 & 0.9448& 0.1589 & 0.6584 & 0.5712 & 0.0138  &  0.8721 & 0.6433 & 0.8768 & 0.6424\\
    & UMATO (ours) & \textbf{\underline{0.3271}} &\textbf{\underline{0.3888}}& \textbf{\underline{0.0529}} & \textbf{\underline{0.1341}} & \textbf{\underline{0.1434}} & \textbf{\underline{0.0014}}  &  0.7884 & \textbf{0.6558} & 0.7887 & \textbf{0.6557}\\
\midrule
    \multirow{7}{*}{\shortstack{Fashion MNIST}} & PCA & 0.9373 &0.2315& \textbf{\underline{0.0255}} & 0.6929 & 0.0454 & \textbf{\underline{0.0006}} &   0.9843 & 0.9117 & 0.9853 & 0.9115\\
    & Isomap & \textbf{\underline{0.9228}} &\textbf{0.2272}& 0.0352 & \textbf{\underline{0.6668}} & \textbf{0.0446} & 0.0010  &  0.9865 & 0.9195 & 0.9872 & 0.9196\\
    & $t$-SNE & 0.9987 & 0.2768& 0.0442 & 0.8079 & 0.0663 & 0.0017  &  0.9899 & \textbf{\underline{0.9949}} & 0.9919 & \textbf{\underline{0.9955}}\\
    & UMAP & 1.0125 &0.2755& 0.0438 & 0.8396 & 0.0641 & 0.0016 &   \textbf{\underline{0.9950}} & 0.9584 & \textbf{\underline{0.9955}} & 0.9584\\
    & TopoAE & 0.9402 &0.2329& \textbf{0.0311} & 0.7301 & 0.0446 & \textbf{0.0008} &   0.9908 & 0.9591 & 0.9913 & 0.9590\\
    & A$t$-SNE & 1.0187 &0.2973& 0.0454 & 0.8389 & 0.0702 & 0.0017   & 0.9826 & \textbf{0.9847} & 0.9849 & \textbf{0.9848}\\
    & UMATO (ours) & \textbf{0.9360} &\textbf{\underline{0.2035}}& 0.0314 & \textbf{0.6852} & \textbf{\underline{0.0342}} & 0.0008  &  \textbf{0.9911} & 0.9500 & \textbf{0.9919} & 0.9502\\
\midrule
    \multirow{7}{*}{MNIST}  & PCA & 1.3237 & 0.4104 & 0.0426 & 1.4981 & 0.1349 & 0.0014 &   0.9573 & 0.7340 & 0.9605 & 0.7342\\
    & Isomap & \textbf{\underline{1.1936}} &\textbf{\underline{0.3358}}& \textbf{0.0382} & \textbf{\underline{1.0361}} & \textbf{\underline{0.0857}} & \textbf{0.0012}  & 0.9743 & 0.7527 & 0.976 & 0.7528\\
    & $t$-SNE & 1.3193 &0.4263& 0.0553 & 1.4964 & 0.1523 & 0.0024 &  \textbf{0.9833} & \textbf{\underline{0.9954}} & \textbf{0.9869} & \textbf{\underline{0.9963}}\\
    & UMAP  & 1.3428 &0.4172 & 0.0588 & 1.5734 & 0.1430 & 0.0026 &   \textbf{\underline{0.9891}} & 0.9547 & \textbf{\underline{0.9907}} & 0.9547\\
    & TopoAE & 1.3038 &0.3686 & \textbf{\underline{0.0366}} & 1.3818 & 0.1048 & \textbf{\underline{0.0011}} &  0.9716 & 0.9429 & 0.9732 & 0.9429\\
    & A$t$-SNE & 1.3312 &0.4328& 0.0466 & 1.5623 & 0.1482 & 0.0018 & 0.9768 & \textbf{0.9765} & 0.9830 & \textbf{0.9777}\\
    & UMATO (ours) & \textbf{1.2738} &\textbf{0.3525}& 0.0414 & \textbf{1.2785} & \textbf{0.1017} & 0.0014  &  0.9792 & 0.8421 & 0.9813 & 0.8422\\
\midrule
    \multirow{7}{*}{\shortstack{Kuzushiji MNIST}} & PCA & 0.3756  &0.4215& 0.0440& 0.1710 & 0.1317 & 0.0014 &   0.9380 & 0.7213 & 0.9420 & 0.7211\\
    & Isomap & 0.5267 &\textbf{0.3458}&  \textbf{0.0379} & 0.2171 & \textbf{0.0906} & \textbf{0.0012} &  0.9573 & 0.7638 & 0.9589 & 0.7635\\
    & $t$-SNE & \textbf{0.2360}&0.4254& 0.0571 & \textbf{0.0483} & 0.1369 & 0.0025 &   0.9843 & \textbf{\underline{0.9688}} & 0.9871 & \textbf{\underline{0.9693}}\\
    & UMAP & \textbf{\underline{0.2126}}&0.3873& 0.0566  & \textbf{\underline{0.0417}} & 0.1148 & 0.0026 &   \textbf{\underline{0.9893}} & 0.9563 & \textbf{\underline{0.9908}} & 0.9564\\
    & TopoAE & 0.3934  &0.3730& \textbf{\underline{0.0373}} & 0.1495 & 0.1027 & \textbf{\underline{0.0011}} & 0.9755 & 0.9442 & 0.9768 & 0.9440\\
    & A$t$-SNE & 0.3029&0.3505&  0.0384 & 0.0807 & 0.0978 & 0.0013 & 0.9786 & \textbf{0.9671} & 0.9824 & \textbf{0.9676}\\
    & UMATO (ours) & 0.3993&\textbf{\underline{0.3231}}& 0.0435& 0.1365 & \textbf{\underline{0.0815}} & 0.0016 & \textbf{0.9865} & 0.8888 & \textbf{0.9881} & 0.8895\\
\bottomrule
\end{tabular}
\vspace{1.5mm}
\caption{
\label{tab:quant} Quantitative results of UMATO and six baseline algorithms. The hyperparameters of the DR algorithms are chosen to minimize $\textrm{KL}_{0.1}$. The best one is bold and underlined, and the runner-up is bold. Only the first four digits are shown for conciseness.}

\end{center}

\vspace{-8mm}
\end{table*}

\section{Quantitative Experiment}

\label{sec:exp}

To evaluate UMATO in preserving global and local structures of HD data, we conducted experiments on one synthetic (Spheres) and three real-world datasets (MNIST \cite{lecun-mnisthandwrittendigit-2010}, Fashion MNIST (FMNIST) \cite{xiao2017-online}, and Kuzushiji MNIST (KMNIST) \cite{clanuwat2018deep}). We compared UMATO with six baseline DR techniques (PCA, Isomap, $t$-SNE, UMAP, TopoAE, and A$t$-SNE) utilizing global (DTM, KL divergence) and local (T\&C, MRREs) quality metrics.

\subsection{Design}

\noindent \textbf{Datasets. } For a synthetic dataset, we used Spheres \cite{Moor19Topological}; it consists of 11 101-dimensional spheres, where ten spheres with relatively small radius of 5 and the number of points of 500 are enclosed by a larger sphere with a radius of 25 and the number of 5,000. A total of 10,000 points form the data. In the case of real-world datasets, we used MNIST, FMNIST, and KMNIST. The datasets represent images of digits, fashion items, and Japanese characters, respectively, and consist of 60,000 rows of 784-dimensional (28 $\times$ 28) vectors.

\noindent \textbf{Metrics. }
To analyze DR techniques in diverse perspectives, we utilized several global and local quality metrics. 
For global metrics, we utilized Distance to a Measure (DTM \cite{chazal2011geometric, chazal2017robust}) and KL divergence. Both assesses how well embeddings capture global structure in terms of density estimation. We selected such density-based metrics to align our experiment with the one of TopoAE \cite{Moor19Topological}. In two functions, the density of each HD point $x$ is defined as $f^X_\sigma (x) := \sum_{y \in X}{\exp{(- \textrm{dist}(x,y)^2 / \sigma)}}$, where $X$ denotes the dataset. The density of corresponding low-dimensional point $y$ is defined as $f^Z_\sigma (z)$, where $Z$ denotes the embedding. Then, DTM is computed as $\sum_{(x,z) \in (X, Z)}  f^X_\sigma (x) - f^Z_\sigma(z)$, while KL divergence is defined as $KL (f^X_\sigma (x) \| f^Z_\sigma (z))$. In accordance with the experiment of TopoAE, we used three $\sigma$ values, 0.01, 0.1, and 1, for both measures.

In the case of local metrics, we used mean relative rank errors (MRREs \cite{lee2007nonlinear}) and Trustworthiness \& Continuity (T\&C \cite{venna2001neighborhood}), which quantifies the quality of embedding based on the preservation of neighborhood structure. While Trustworthiness and MRRE$_Z$ evaluate the extent to which neighbors in the embedding are also neighbors in the HD space, Continuity and MRRE$_X$ evaluate the opposite. 
These two metrics are selected because they have been widely used in DR literature \cite{Moor19Topological, jeon2022measuring, gracia2014methodology}. They require a hyperparameter $k$, the number of nearest neighbors; we used $k=5$ throughout our experiments, aligned with the experiment of TopoAE \cite{Moor19Topological}. The results with $k=10$ and $k=15$ are depicted in Appendix C.

\begin{figure}[t]
    \centering
    \includegraphics[width=\linewidth]{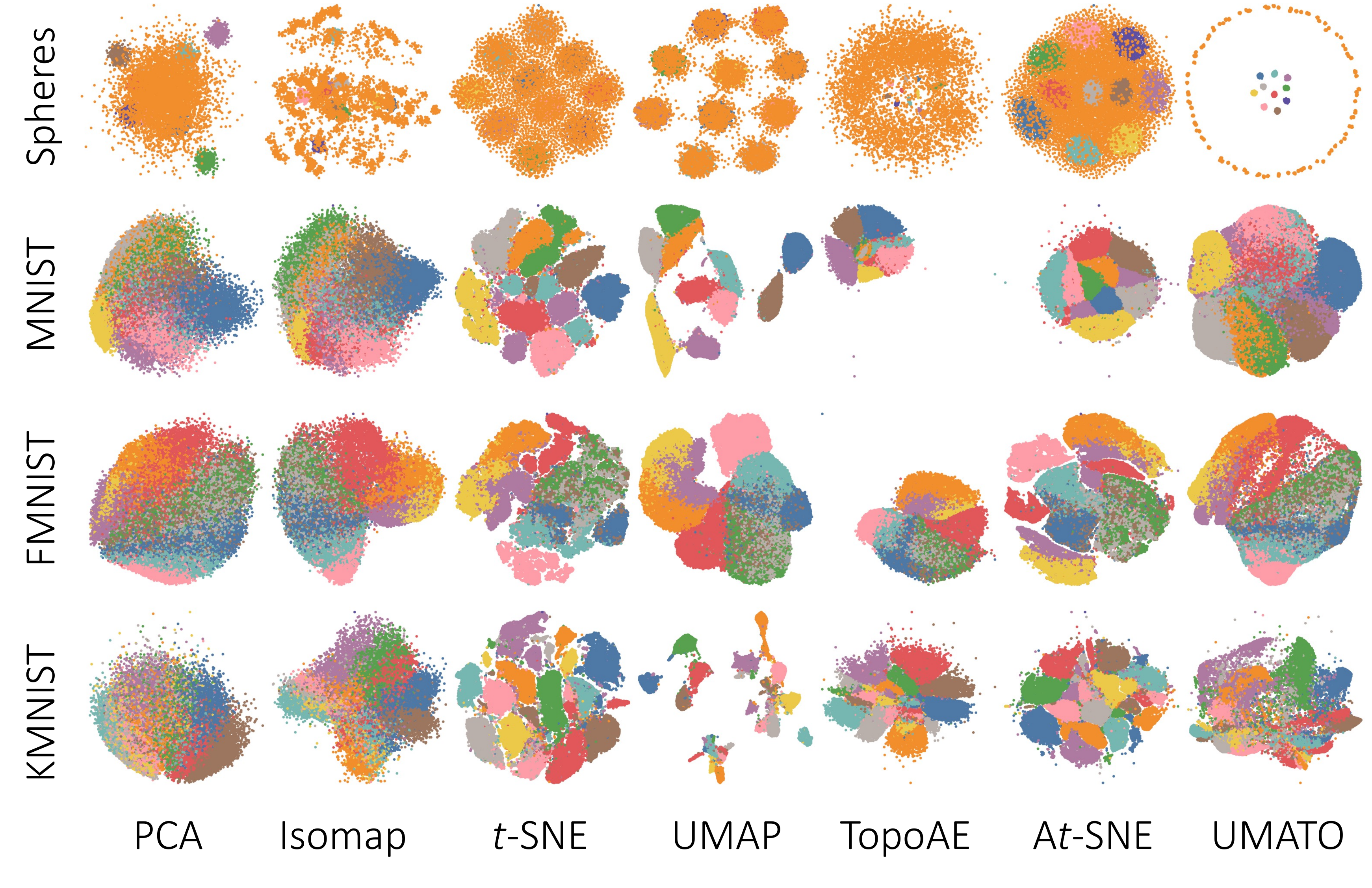}
    \vspace{-7mm}
    \caption{2D embeddings of UMATO and six competitors. Overall, UMATO surpassed other techniques in preserving global structure while showing comparable performance in capturing local structure. A high-resolution version of this figure is provided in Appendix J.  }
    \label{fig:embedding}
    \vspace{-5.5mm}
\end{figure}

\noindent \textbf{Baseline DR techniques. } 
For competitors, we used the most widely used techniques plus the ones that focus on global structures (\autoref{sec:gldr}). For the former, PCA, $t$-SNE, and UMAP were picked, and for the latter, Isomap, A$t$-SNE, and TopoAE, were picked.  
To initialize an embedding, we used PCA for $t$-SNE, following the recommendation in previous work \cite{linderman2019fast}, and spectral embedding for UMAP, which is the default.
The embeddings are depicted in \autoref{fig:embedding}. The hyperparameters of the techniques are set to minimize KL$_{0.1}$.
Appendix B describes our hyperparameter setting in more detail.

\subsection{Results}

\autoref{tab:quant} displays the quantitative evaluation results.
In terms of global metrics, UMATO surpassed all other techniques for Spheres and exceeded all techniques except Isomap for MNIST and FMNIST. Although UMATO did not indisputably show the best performance for KMNIST, it still surpassed other techniques if $\sigma=0.1$, while there was no winner for $\sigma=0.01$ and 1. TopoAE and Isomap also showed good performance in preserving global metrics overall. Although A$t$-SNE uitlizes landmark based optimization, which is similar to UMATO, it showed competitive scores to neither UMATO nor other global techniques.
For local metrics, $t$-SNE and UMAP were common winners. However, UMATO showed competitive performance in terms of  $\textrm{MRRE}_X$ and continuity for FMNIST and KMNIST and in terms of MRRE$_Z$ and Trustworthiness for Spheres.  

\subsection{Discussions}

The results clearly show the benefit of UMATO in preserving the global structure. The shape of embeddings also supports such a claim. If we look at the first row of \autoref{fig:embedding}, we can see that the outer sphere encircles the inner spheres in a circular form in UMATO's embedding, which is the most intuitive to understand the global relationship among different classes. However, all other embeddings did not show a clear separation between the outer and inner spheres. 

Combining results from both global and local metrics, we can conclude that UMATO achieved high performance in capturing global structure at a slight loss for local structure preservation. Such a result aligns well with the design of UMATO; hubs help UMATO capture global structure in the first phase but work as a constraint for local optimization of eNN. Still, the fact that UMATO showed competitive performance in preserving local structures to $t$-SNE and UMAP clearly demonstrates its merit in analyzing HD data.

Refer to Appendix I for the results using synthetic datasets (e.g., Swiss Roll) that support our claim about UMATO's capability.

%% file: sections/05-additional-experiments.tex
\section{Additional Experiments}

We conducted additional experiments to understand UMATO's features better and verify its benefit. At first, we \rev{compared the runtime of UMATO against previous nonlinear DR techniques (Appendix G).}
We also verified that UMATO avoids UMAP's problem of clusters' becoming dispersed as the number of epochs increases (Appendix D), and revealed that UMATO outperforms UMAP and $t$-SNE in terms of stability over subsampling (Appendix E) and robustness over diverse initializations (Appendix F). Finally, we discovered that utilizing more than two phases failed to improve the embedding results while consuming more computation time (Appendix H). 

%% file: sections/06-conclusion.tex
\section{Conclusion}

We present a two-phase DR algorithm called UMATO that can effectively preserve the global and local properties of HD data.
In our experiments with diverse datasets, we have proven that UMATO outperforms previous widely used baselines (e.g., $t$-SNE and UMAP) in capturing global structures while showing competitive performance in preserving local structures.
In the future, we plan to optimize and accelerate UMATO in a heterogeneous system (e.g., GPU), as in previous attempts with other DR techniques \cite{pezzotti2019gpgpu, nolet2020bringing}.
Qualitatively analyzing UMATO's effect in analyzing HD data, such as visual cluster analysis \cite{xia2021revisiting, jeon2022distortion, cavallo2018clustrophile}, will also be an interesting direction.